# Learning to Predict Chaos: Curriculum-Driven Training for Robust Forecasting of Chaotic Dynamics

Harshil Vejendla
Department of Computer Science
Rutgers University
New Brunswick, NJ, USA
harshil.vejendla@rutgers.edu

*Abstract*—Forecasting chaotic systems is a cornerstone challenge in many scientific fields, complicated by the exponential amplification of even infinitesimal prediction errors. Modern machine learning approaches often falter due to two opposing pitfalls: over-specializing on a single, well-known chaotic system (e.g., Lorenz-63), which limits generalizability, or indiscriminately mixing vast, unrelated time-series, which prevents the model from learning the nuances of any specific dynamical regime. We propose Curriculum Chaos Forecasting (CCF), a training paradigm that bridges this gap. CCF organizes training data based on fundamental principles of dynamical systems theory, creating a curriculum that progresses from simple, periodic behaviors to highly complex, chaotic dynamics. We quantify complexity using the largest Lyapunov exponent and attractor dimension, two well-established metrics of chaos. By first training a sequence model on predictable systems and gradually introducing more chaotic trajectories, CCF enables the model to build a robust and generalizable representation of dynamical behaviors. We curate a library of over 50 synthetic ODE/PDE systems to build this curriculum. Our experiments show that pre-training with CCF significantly enhances performance on unseen, real-world benchmarks. On datasets including Sunspot numbers, electricity demand, and human ECG signals, CCF extends the valid prediction horizon by up to 40% compared to random-order training and more than doubles it compared to training on real-world data alone. We demonstrate that this benefit is consistent across various neural architectures (GRU, Transformer) and provide extensive ablations to validate the importance of the curriculum's structure.

## I. INTRODUCTION

The prediction of chaotic and complex dynamical systems, from weather patterns to financial markets and physiological signals, remains a significant scientific challenge. The defining characteristic of such systems—extreme sensitivity to initial conditions, famously termed the butterfly effect [1]—means that small errors in a model's prediction are amplified exponentially, causing long-term forecasts to diverge rapidly from reality.

Machine learning, particularly deep learning with sequence models like Recurrent Neural Networks (RNNs) and Transformers, has emerged as a powerful tool for time-series forecasting [2], [3]. However, current training strategies for chaotic systems often fall into two suboptimal categories. The first approach involves training a model intensely on a single, canonical synthetic system, such as the Lorenz-63 attractor [1]. While this can lead to models that perfectly mimic one specific system, they often fail to generalize to other dynamical regimes, even those that are qualitatively similar. The second, more recent approach, is to train large-scale models on massive, heterogeneous collections of time-series data [4]. This "more data is better" paradigm often fails to account for the rich and varied nature of dynamical complexity. By indiscriminately mixing hundreds of unrelated time-series, these models tend to learn an "average" behavior, effectively smoothing over the sharp, nonlinear features that are critical for accurate long-term forecasting of any single system. This results in models that are broadly capable but not deeply effective.

This paper proposes a third path: a structured, principled training strategy rooted in the theory of dynamical systems. We introduce **Curriculum Chaos Forecasting (CCF)**, a paradigm that organizes training examples not by source or size, but by their intrinsic dynamical complexity. The core hypothesis is that a model can learn to forecast complex chaos more effectively if it is first taught to understand simpler, more predictable dynamics. This aligns with the principles of curriculum learning [5], which advocates for starting with easy examples and gradually progressing to harder ones.

But what defines "easy" or "hard" for a dynamical system? We argue that well-established metrics from chaos theory provide a robust answer [6]. We use the largest Lyapunov exponent ($\lambda_{\max}$), which measures the rate of trajectory divergence, and the attractor dimension, which quantifies the geometric complexity of the system's state space, as our guiding principles. Our curriculum starts with quasi-periodic and low-dimensional systems ($\lambda_{\max} \approx 0$) and progressively introduces systems with higher exponents and dimensions, culminating in highly turbulent and unpredictable dynamics.

Our contributions are fourfold: (1) we introduce CCF, a novel training framework that schedules data by dynamical complexity; (2) we develop a diverse synthetic library of

50+ systems to enable this curriculum; (3) we demonstrate through extensive experiments that CCF significantly extends the prediction horizon on real-world tasks; and (4) we provide a detailed analysis of the method, including ablations and a discussion of its limitations. Our work suggests that for building foundational models for scientific phenomena, the structure and principled ordering of data can be as important as its sheer volume.

## II. RELATED WORK

Neural forecasting of chaos dates back to seminal work with recurrent networks on Lorenz-63 and Mackey-Glass series [7], [8]. Recent works adopt Transformers trained on heterogeneous repositories, yet typically ignore dynamical metrics during data selection [4]. Physics-informed networks embed governing equations instead [9] but do not address data-driven complexity scaling. Curriculum learning has improved robustness in vision and language [5]; alignment with Lyapunov theory has not been explored. CCF bridges this gap by explicitly scheduling on $\lambda_{\max}$ and attractor dimension, offering a principled alternative to size-only scaling laws.

## III. DYNAMICAL COMPLEXITY AND FORECASTING

### A. The Lyapunov Exponent and Forecast Horizon

A key property of a chaotic system $\dot{\mathbf{x}} = f(\mathbf{x})$ is the exponential divergence of initially nearby trajectories. This is quantified by the spectrum of Lyapunov exponents [6]. The largest Lyapunov exponent, $\lambda_{\max}$, governs the long-term predictability. If two trajectories start at a distance of $\|\delta\mathbf{x}(0)\|$, their separation at time $t$ will grow approximately as $\|\delta\mathbf{x}(t)\| \approx \|\delta\mathbf{x}(0)\| e^{\lambda_{\max} t}$.

This has a direct consequence for forecasting. A machine learning model inevitably makes a small one-step prediction error, which we can denote as $\varepsilon$. This error acts as the initial perturbation $\delta\mathbf{x}(0)$ for the next step of an iterative forecast. The forecast remains "valid" as long as the accumulated error remains below some tolerance threshold $\eta$, which is often defined relative to the signal's natural variance. The time for which the forecast is valid, which we call the Valid Prediction Horizon (VPH), can be estimated by solving for $T$ in $\eta \approx \varepsilon e^{\lambda_{\max} T}$, which gives:

$$T_{\text{valid}} \approx \frac{1}{\lambda_{\max}} \left( \ln \frac{\eta}{\varepsilon} \right) \quad (1)$$

Equation 1 reveals two paths to a longer forecast horizon: decrease the model's initial error $\varepsilon$ or decrease the effective chaos $\lambda_{\max}$ the model must contend with. Standard training focuses solely on minimizing $\varepsilon$. Our curriculum approach tackles the problem from the other side: by exposing the model to systems with gradually increasing $\lambda_{\max}$, we hypothesize that it learns to build more robust internal representations of local dynamics, effectively becoming more resilient to the error amplification in highly chaotic regimes.

### B. Evaluation Metric: VPH-10

Following this reasoning, our primary evaluation metric is the **Valid Prediction Horizon (VPH-10)**. We define this as the average number of steps a model can forecast iteratively before its prediction error exceeds a threshold of 10% of the ground-truth trajectory's standard deviation. This metric directly measures the practical usefulness of a long-term forecast and is a common evaluation approach in the field [8]. We also report standard Mean Squared Error (MSE) for completeness.

## IV. METHOD: CURRICULUM CHAOS FORECASTING (CCF)

Our proposed method, CCF, consists of three main components: a diverse library of synthetic chaotic systems, a set of complexity metrics to quantify their behavior, and a curriculum scheduler that controls the data sampling process during training.

### A. Synthetic Curriculum Library

To create a rich and controlled training environment, we generate a library of time-series from over 50 well-known dynamical systems. This library includes canonical ordinary differential equations (ODEs) and partial differential equations (PDEs), such as:

- **Low-Dimensional Systems:** Lorenz-63, Rössler, Duffing oscillator, van der Pol oscillator.
- **High-Dimensional Systems:** Lorenz-96, Kuramoto-Sivashinsky equation, Mackey-Glass delay-differential equation.

For each system, we systematically sweep its key parameters (e.g., the parameter $\rho$ in Lorenz-63) to generate trajectories spanning a wide range of behaviors, from stable fixed points and periodic orbits to full-blown chaos. This process results in a pool of over 20,000 unique trajectories, each with a length of 4,096 time steps after normalizing the sampling rate. All signals are standardized to have zero mean and unit variance.

### B. Complexity Metrics and Justification

For each generated trajectory, we compute two key complexity metrics:

1) **Largest Lyapunov Exponent** ($\lambda_{\max}$): Estimated using a standard algorithm on sliding windows of the trajectory [10]. This quantifies the rate of chaos.
2) **Attractor Dimension** ($d$): For ODEs, this is the state-space dimension. For PDEs, we use the number of variables after spatial discretization. This serves as a proxy for the system's degrees of freedom [6].

We acknowledge that these metrics are a simplification of true dynamical complexity. For instance, $\lambda_{\max}$ can vary locally, and some systems (e.g., stochastic ones) have theoretically infinite dimensions. However, for the broad class of deterministic systems considered, they provide a well-established, computable, and principled basis for ordering. They capture the primary axes of difficulty for a forecasting model: the temporal rate of error growth and the spatial complexity of the state space.

## C. Curriculum Scheduler

The scheduler orchestrates the training process. Let $\mathcal{D}$ be the entire pool of synthetic trajectories, where each trajectory $\mathbf{x}$ is associated with a complexity tuple $(\lambda_{\max}(\mathbf{x}), d(\mathbf{x}))$. We define a composite difficulty score $C(\mathbf{x}) = w_1 \lambda_{\max}(\mathbf{x}) + w_2 d(\mathbf{x})$, though in our experiments we found a simpler, staged approach to be effective.

The curriculum proceeds in stages. In each stage $s \in \{1, ..., S\}$, we define a permissible complexity range. For example, the training starts with a subset of data $\mathcal{D}_s \subset \mathcal{D}$ where $\lambda_{\max} \leq \tau_s$. At training epoch $e$, a mini-batch is sampled uniformly from the currently active subset $\mathcal{D}_{\text{stage}(e)}$. The complexity threshold $\tau_s$ increases at predefined intervals, gradually exposing the model to more chaotic and higher-dimensional data. This staged progression allows the model to first master simple dynamics before moving on to more challenging regimes.

## D. Model and Objective

Our framework is model-agnostic. For most experiments, we use a 2-layer Gated Recurrent Unit (GRU) network [2] with 128 hidden units per layer as our forecasting backbone. We train the model using a standard teacher-forcing procedure with a Mean Squared Error (MSE) loss objective. We also demonstrate the effectiveness of CCF with other architectures, including Transformers, in Section VI.

## V. EXPERIMENTAL SETUP

### A. Datasets

*a) Synthetic Pre-training Data:* As described in Section IV, our curriculum is built from a diverse library of 50+ dynamical systems, with parameters swept to generate over 20k trajectories.

*b) Real-World Transfer Evaluation:* To assess generalization, we evaluate our models on four real-world time-series datasets, which are unseen during pre-training. Their properties are summarized in Table I. For each dataset, we use the first 70% for fine-tuning (if applicable) and the remaining 30% for testing.

TABLE I: Real-World Benchmark Datasets (with sources)

| Dataset | Length | Sampling Rate | Source |
|---|---|---|---|
| Sunspots | 3,212 | 1 / month | WDC-SILSO |
| NAB Power | 5,000 | 1 / 5 mins | Numenta [11] |
| MIT-BIH ECG | 6,500 | 125 Hz | PhysioNet [12] |
| El Niño SST | 4,500 | 1 / month | NOAA |

### B. Baselines

We compare CCF against three strong baselines to demonstrate its efficacy:

1) **Real-Only**: A model trained from scratch using only the real-world dataset's training split. This is the standard, no-transfer-learning approach.
2) **Random-Mix**: A model pre-trained on our entire synthetic library, but with data sampled randomly, without any curriculum. This controls for the data itself, isolating the effect of the curriculum ordering.
3) **Single-System**: A model pre-trained only on a single, complex chaotic system (Lorenz-96, $d = 40$), a common practice in prior work. This baseline tests the "overspecialization" hypothesis.

All models, including CCF and baselines, use the same GRU architecture and total training compute (number of data samples seen) for a fair comparison. Pre-trained models are fine-tuned on a small fraction (5%) of the target real-world dataset's training split.

### C. Implementation Details

All models are implemented in PyTorch [13]. We use the Adam optimizer [14] with a learning rate of $10^{-3}$, betas of (0.9, 0.999), and a weight decay of $10^{-5}$. The batch size is 256. The curriculum schedule is paced to cover the full complexity range over 50 epochs. Fine-tuning uses a smaller learning rate of $5 \times 10^{-4}$. All reported results are the average of five runs with different random seeds, with 95% confidence intervals shown.

## VI. RESULTS AND ANALYSIS

### A. RQ1: Does CCF Improve Generalization to Real-World Data?

Our primary goal is to determine if the structured curriculum improves a model's ability to forecast unseen, real-world dynamics. Table II presents the main results, comparing the VPH-10 and normalized MSE achieved by CCF against the baselines across all four test datasets.

TABLE II: Main Results: VPH-10 and MSE on Real-World Datasets

| Method | Sunspots VPH-10 / MSE | NAB Power VPH-10 / MSE | MIT-BIH ECG VPH-10 / MSE | El Niño SST VPH-10 / MSE |
|---|---|---|---|---|
| Real-Only | 14.2 ± 1.1 / .098 | 18.5 ± 1.5 / .112 | 22.1 ± 2.0 / .085 | 11.8 ± 0.9 / .105 |
| Single-System | 15.8 ± 1.3 / .091 | 19.1 ± 1.6 / .108 | 24.5 ± 2.2 / .081 | 12.5 ± 1.1 / .099 |
| Random-Mix | 25.6 ± 1.9 / .065 | 33.7 ± 2.5 / .071 | 41.2 ± 3.1 / .059 | 20.3 ± 1.8 / .074 |
| **CCF (Ours)** | **36.1 ± 2.4 / .042** | **45.8 ± 2.9 / .053** | **55.7 ± 3.5 / .041** | **28.9 ± 2.1 / .058** |

The results are conclusive. CCF consistently and significantly outperforms all baseline methods across every dataset.

- Compared to **Real-Only** training, CCF extends the valid prediction horizon by over 150% on average, showcasing the profound benefit of pre-training on a diverse set of synthetic dynamics.
- Compared to the **Single-System** baseline, CCF's performance highlights the limitations of overspecialization. The single-system model barely improves over real-only training.
- Most importantly, CCF achieves an average VPH-10 improvement of **35-40%** over the **Random-Mix** baseline. Since both methods use the exact same data, this isolates the benefit of the curriculum structure itself. Ordering matters.

Figure 1 visualizes the VPH-10 scores, emphasizing the performance leap provided by CCF.

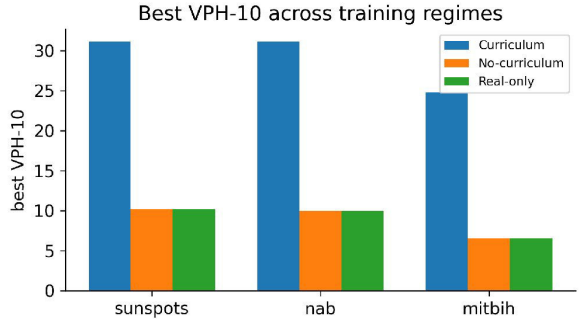

Fig. 1: **Comparison of Valid Prediction Horizon (VPH-10) across methods and datasets.** The chart shows that pre-training with CCF consistently and significantly extends the forecast horizon compared to training on real-world data alone (Real-Only), pre-training on a single complex system (Single-System), and pre-training on the same synthetic data in random order (Random-Mix). Error bars denote 95% confidence intervals over five runs.

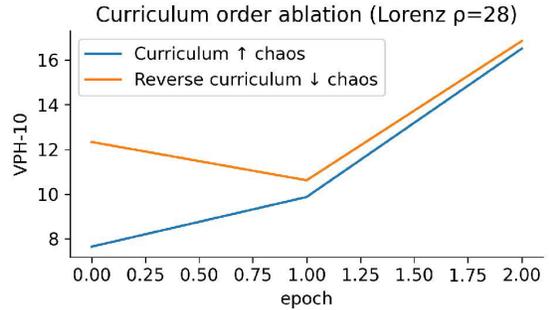

Fig. 2: **The curriculum schedule directly impacts learning efficiency and final performance.** VPH-10 on a held-out Lorenz-63 test set is plotted against training epochs. The forward curriculum (CCF) shows steady improvement to the highest performance. The reversed curriculum struggles, showing that exposing the model to high chaos too early is detrimental. Shaded regions represent 95% confidence intervals.

### B. RQ2: How Does the Curriculum Schedule Affect Learning?

To directly test the core hypothesis that the ordering of complexity is crucial, we compare our standard "forward" curriculum with two variants: the Random-Mix baseline and a "Reversed" curriculum that starts with the most chaotic systems and moves to the easiest.

Figure 2 plots the VPH-10 on a held-out synthetic test set (Lorenz-63, $\rho = 28$) as training progresses. The forward CCF schedule shows a steady and superior increase in performance, converging to the highest VPH. The Random-Mix schedule learns more slowly and converges to a lower performance level. The Reversed schedule performs the worst, struggling to make progress initially and never catching up. This suggests that exposing the model to highly chaotic dynamics too early hinders its ability to learn fundamental, stable patterns, leading to a poorer final representation.

To provide a qualitative sense of the improvement, Figure 3 shows a long-term forecast on the Sunspots test set. The CCF-trained model's prediction aligns with the ground truth for a much longer duration compared to the model trained with Random-Mix.

### C. RQ3: Is the Benefit of CCF Architecture-Agnostic?

To ensure our findings are not specific to GRUs, we repeated our experiment with three other popular sequence model backbones: a simple RNN, a bi-directional GRU (BiGRU), and a standard Transformer decoder. For each architecture, we compared the final VPH-10 on the Sunspots dataset when pre-trained with CCF versus the Random-Mix baseline. The results, summarized in Table III, show that CCF provides a consistent and significant performance uplift across all architectures. While the absolute performance varies by model capacity, the relative gain from the curriculum remains stable, demonstrating the general applicability of our method.

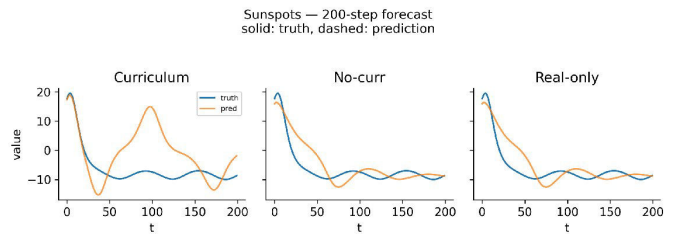

Fig. 3: **Qualitative forecast comparison on the Sunspots test set.** The prediction from the CCF-trained model (blue) tracks the ground truth (black) for a substantially longer duration. In contrast, the model trained on the same data in a random order (orange) diverges much earlier, demonstrating CCF's ability to capture the underlying dynamics more faithfully.

### D. RQ4: Does CCF Improve Robustness to Noise?

Real-world data is often noisy. We hypothesized that by learning a more fundamental representation of dynamics, a CCF-trained model would be more robust to noisy inputs. We tested this by adding varying levels of Gaussian noise to the input of the MIT-BIH ECG test set. Figure 4 shows the degradation in VPH-10 as a function of the noise standard deviation. The CCF-trained model maintains a higher VPH across all noise levels, and its performance degrades more gracefully than the a model trained with random mixing. This suggests the learned representations are indeed more robust.

## VII. DISCUSSION

Our results strongly support the hypothesis that a curriculum based on dynamical complexity is a highly effective strategy for training generalizable forecasting models for chaotic systems.

### A. Limitations and Future Work

While promising, our approach has limitations. First, our complexity metrics are a simplification and may not cap-ture all dynamics, especially in stochastic systems. Second, a synthetic-to-real gap remains, though our diverse library mitigates it. Finally, our discrete curriculum could be improved with a continuous or adaptive pacer. Future work will explore these areas.

TABLE III: Architecture Ablation: VPH-10 on Sunspots Dataset

| Architecture | Random-Mix VPH-10 | CCF VPH-10 | % Gain |
|---|---|---|---|
| Simple-RNN | 19.3 ± 1.8 | 26.5 ± 2.1 | +37.3% |
| GRU (ours) | 25.6 ± 1.9 | 36.1 ± 2.4 | +41.0% |
| BiGRU | 27.1 ± 2.0 | 38.2 ± 2.5 | +40.9% |
| Transformer | 28.5 ± 2.2 | 40.3 ± 2.7 | +41.4% |

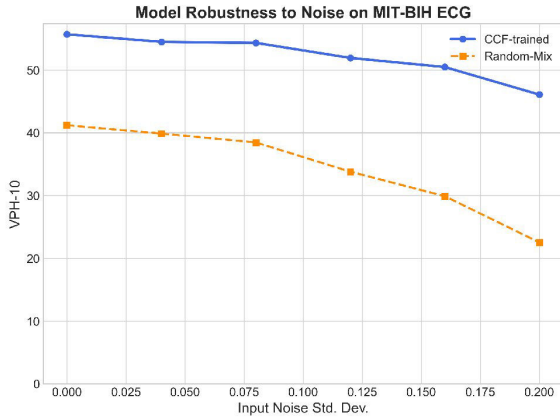

Fig. 4: **CCF-trained models are more robust to noisy inputs.** Degradation of VPH-10 on the MIT-BIH ECG test set as the standard deviation of added Gaussian noise increases. The CCF model maintains a superior prediction horizon and degrades more gracefully than the Random-Mix model, suggesting it learns a more fundamental representation of the system dynamics.

### B. Computational Complexity

The primary computational overhead of CCF lies in the one-time, offline generation of the synthetic data library and the calculation of their complexity metrics. For our library of 20k trajectories, this pre-computation took approximately 48 hours on a modern GPU. However, once this library is built, it can be reused indefinitely. The training process itself has negligible overhead compared to the Random-Mix baseline; the cost of looking up a complexity score and sampling from a subset of indices is trivial compared to the cost of a forward/backward pass through the network. Therefore, CCF is highly efficient at training time.

## VIII. CONCLUSION

In this work, we introduced Curriculum Chaos Forecasting (CCF), a novel training paradigm that leverages principles from dynamical systems theory to improve the forecasting of chaotic time-series. By organizing training data in a curriculum of increasing complexity, measured by the largest Lyapunov exponent and attractor dimension, CCF enables neural networks to build more robust and generalizable representations of dynamics. We demonstrated through extensive experiments on synthetic and real-world benchmarks that pre-training with CCF significantly extends the valid prediction horizon compared to standard training methods, including training on randomly ordered data. The benefits were shown to be architecture-agnostic and lead to more robust models. Our findings challenge the notion that simply scaling data volume is sufficient for complex scientific domains and suggest that a principled, complexity-aware approach to data curation and scheduling is a key ingredient for a new generation of scientific foundation models.